\documentclass{article}

\usepackage[preprint]{neurips_2025}

\usepackage[utf8]{inputenc} % allow utf-8 input
\usepackage[T1]{fontenc}    % use 8-bit T1 fonts
\usepackage{hyperref}       % hyperlinks
\usepackage{url}            % simple URL typesetting
\usepackage{booktabs}       % professional-quality tables
\usepackage{amsfonts}       % blackboard math symbols
\usepackage{nicefrac}       % compact symbols for 1/2, etc.
\usepackage{microtype}      % microtypography
\usepackage{xcolor}         % colors

\usepackage[ruled,vlined,linesnumbered]{algorithm2e}
\usepackage{adjustbox}
\usepackage{enumitem}
\usepackage{graphicx}
\usepackage{multirow}
\usepackage{lineno}
\usepackage{subcaption}
\usepackage{wrapfig}
\usepackage[most]{tcolorbox}
\usepackage{soul}
\usepackage{placeins}

\makeatletter
\patchcmd{\@algocf@start}% <cmd>
  {-1.5em}% <search>
  {0pt}% <replace>
  {}{}% <success><failure>
\makeatother

\title{LLM-Integrated Bayesian State Space Models\\ for Multimodal Time-Series Forecasting}

\author{%
  Sungjun Cho, Changho Shin, Suenggwan Jo,\\
  \textbf{Xinya Yan, Shourjo Aditya Chaudhuri, Frederic Sala}\\
  Department of Computer Sciences \\
  University of Wisconsin-Madison \\
  \texttt{\{cho266, cshin23, sjo32, xyan89, sachaudhuri, fsala\}@wisc.edu} \\
}

\newcommand{\ourmethod}{LBS}

\newcommand{\hlc}[2][yellow]{{%
    \colorlet{foo}{#1}%
    \sethlcolor{foo}\hl{#2}}%
}

\usepackage{amsmath,amsfonts,bm}

\def\eqref#1{equation~\ref{#1}}
\def\1{\bm{1}}

\def\vh{{\bm{h}}}

\def\vs{{\bm{s}}}

\def\vx{{\bm{x}}}
\def\vy{{\bm{y}}}

\DeclareMathAlphabet{\mathsfit}{\encodingdefault}{\sfdefault}{m}{sl}
\SetMathAlphabet{\mathsfit}{bold}{\encodingdefault}{\sfdefault}{bx}{n}

\def\gD{{\mathcal{D}}}

\def\gL{{\mathcal{L}}}

\newcommand{\E}{\mathbb{E}}

\newcommand{\R}{\mathbb{R}}

\begin{document}

\maketitle

\begin{abstract}

Forecasting in the real world requires integrating structured time-series data with unstructured textual information, but existing methods are architecturally limited by fixed input/output horizons and are unable to model or quantify uncertainty. We address this challenge by introducing LLM-integrated Bayesian State space models (LBS), a novel probabilistic framework for multimodal temporal forecasting. At a high level, \ourmethod{} consists of two components: (1) a state space model (SSM) backbone that captures the temporal dynamics of latent states from which both numerical and textual observations are generated and (2) a pretrained large language model (LLM) that is adapted to encode textual inputs for posterior state estimation and decode textual forecasts consistent with the latent trajectory. This design enables flexible lookback and forecast windows, principled uncertainty quantification, and improved temporal generalization thanks to the well-suited inductive bias of SSMs toward modeling dynamical systems. Experiments on the TextTimeCorpus benchmark demonstrate that \ourmethod{} improves the previous state-of-the-art by 13.20\% while providing human-readable summaries of each forecast. \textbf{Our work is the first to unify LLMs and SSMs for joint numerical and textual prediction, offering a novel foundation for multimodal temporal reasoning}.
\end{abstract}

\section{Introduction}\label{sec:introduction}

Time-series forecasting, a core machine learning task, is traditionally centered on predicting future numerical values from past data~\cite{(patchtst)nie2022time}. However, in many real-world domains, contextual information expressed in natural language---such as clinical notes, financial reports, or weather descriptions---plays a critical role in forecasting. This complementary modality can offer valuable signals that cannot be fully extracted from numeric data alone~\cite{liu2024time,kim2024multi}. Similarly, generating textual forecasts alongside numerical predictions can improve interpretability. This is particularly useful in high-stakes decision-making scenarios. These opportunities motivate the development of models that not only forecast from multimodal inputs, but also communicate their predictions through natural language, augmenting quantitative accuracy with qualitative explanations. 

Probabilistic state space models (SSMs) offer compelling advantages for time-series forecasting: they can quantify uncertainty and support flexible, variable-length prediction horizons. For this reason, it is intuitive to seek to integrate a probabilistic SSM with a pretrained large language model (LLM), enabling forecasting from and into both numeric and textual modalities. However, this integration presents two fundamental  challenges: \textbf{(C1) Text-conditioned posterior state estimation:} How can we update the latent state of the SSM using a pretrained LLM and textual observations? \textbf{(C2) Latent state-conditioned text generation:} How can we adapt the LLM to generate accurate, temporally grounded textual forecasts conditioned on latent state trajectories?
%These challenges lie at the core of multimodal temporal modeling but remain largely unexplored.
%Probabilistic state space models (SSMs) offer key advantages for time-series forecasting by modeling uncertainty and supporting variable-length input and output horizons. A natural next step is to integrate a probabilistic SSM with a pretrained large language model (LLM) to enable both textual inputs and outputs. However, this integration introduces two fundamental and underexplored challenges: \textbf{(C1) Text-conditioned posterior state estimation:} How can we update the latent state of the SSM using textual observations processed by an LLM? \textbf{(C2) Latent state-conditioned text generation:} How can we adapt the LLM to generate accurate and temporally grounded textual forecasts conditioned on latent state trajectories? These are problems that have not yet been studied in the context of multimodal temporal forecasting.

\begin{figure}[!t]
    \centering
    \resizebox{\textwidth}{!}
    {
    \includegraphics[width=\linewidth]{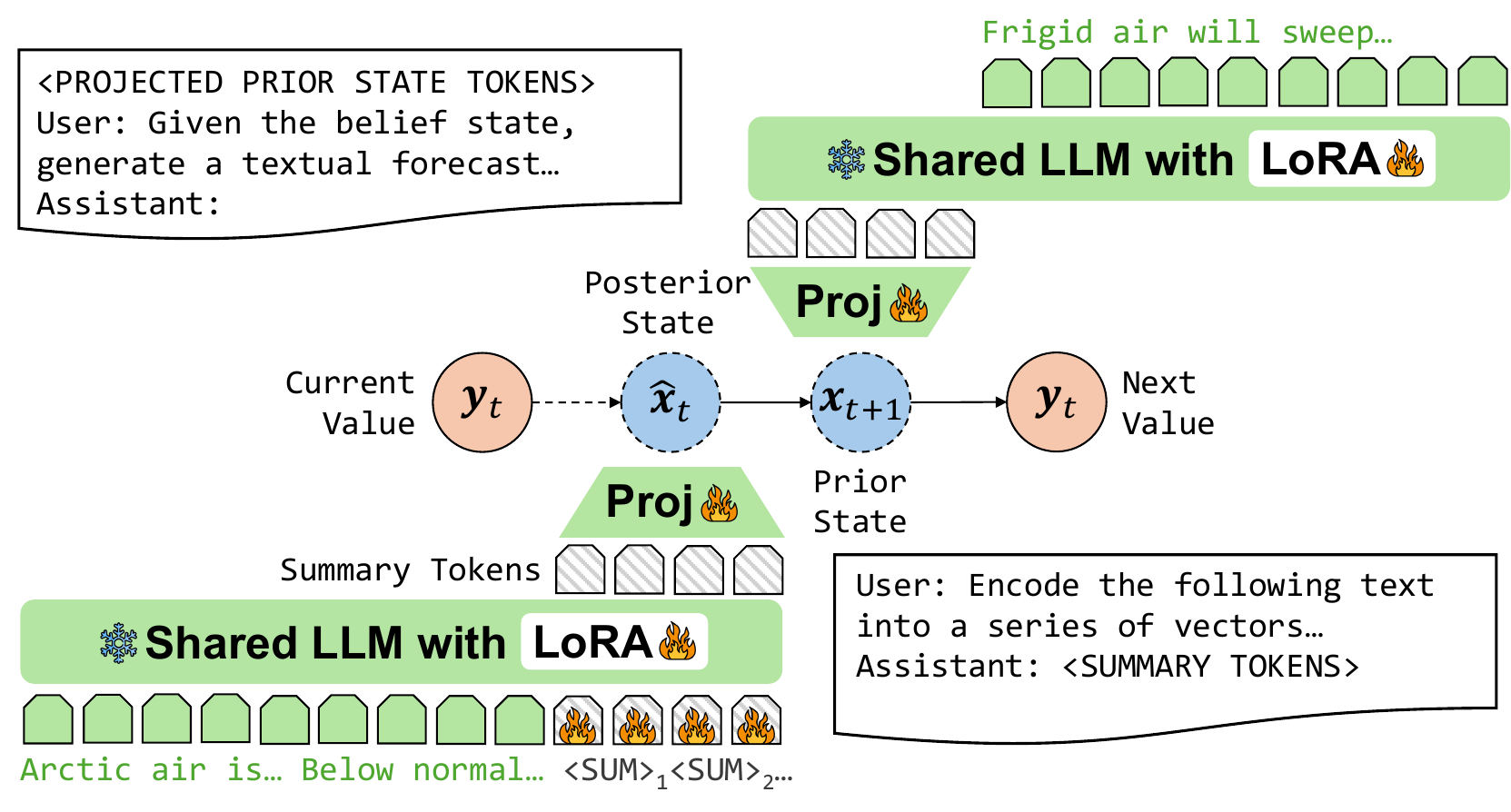}
    }
    \vspace{-5px}
    \caption{An illustration of the \ourmethod{} architecture under for a  single-step forecasting scenario. \textbf{Bottom:} To enable variational inference with text, the LLM is tuned to summarize the context into a set of summarization tokens, which are then used together with the target value to obtain the posterior state distribution. \textbf{Top:} Conditioned on the state forecast, the shared LLM is trained to generate its corresponding text. When using instruction-tuned LLMs, both steps are accompanied by prompt templates in order to preserve its capabilities.
    }
    \label{fig:autoencoder}
    \vspace{-25px}
\end{figure}

To address these challenges, we propose the \textbf{LLM-integrated Bayesian State Space Model (\ourmethod{})}, a novel architecture that unifies a probabilistic SSM with a pretrained LLM for joint numeric and textual forecasting (see \autoref{fig:autoencoder}). For \textbf{(C1)}, we adapt the LLM to summarize and compress textual inputs into a sequence of summary tokens, which are then together projected into the low-dimensional latent state space for deep Bayesian filtering. For \textbf{(C2)}, we leverage the LLM’s in-context generation capabilities by conditioning it on latent state trajectories---treated as non-textual context, akin to images or videos---enabling the model to generate temporally coherent, state-grounded textual forecasts. To our knowledge, \ourmethod{} is the first framework to tightly couple probabilistic SSMs and LLMs to generate temporally coherent forecasts across both numeric and textual modalities. 

We evaluate \ourmethod{} on the \textsc{TimeTextCorpus (TTC)}, a multimodal benchmark spanning  climate and clinical domains. Our model consistently outperforms both unimodal and multimodal baselines, \textit{\textbf{improving numeric forecasting accuracy by an average of 13.20\%}} while generating coherent and contextually relevant textual predictions. Qualitative analyses show further benefits of the Bayesian SSM backbone, including its \textit{\textbf{ability to encode trends and seasonality in low-dimensional latent states and provide principled uncertainty estimates}}. Finally, we explore LLM scaling by substituting various models from the Qwen2.5~\cite{yang2024qwen2} and LLaMA3~\cite{grattafiori2024llama} families, finding that larger LLMs do not always improve performance, but \textit{\textbf{incorporating textual context enhances robustness especially for long-horizon forecasts}}.

The key contributions of our work are the following:
\begin{itemize}[leftmargin=15px,noitemsep, topsep=0.5pt]
    \item We present \ourmethod, a novel architecture for multimodal time-series forecasting that integrates a Bayesian state space model with pretrained LLMs to enable conditioning posterior estimates on textual inputs as well as text generation.
    \item We incorporate techniques from the LLM-based context compression and multimodal instruction tuning literature towards enabling \ourmethod{} to perform posterior state estimation conditioned on text and state-conditioned text generation.
    \item We verify the empirical performance of \ourmethod{} on the multimodal forecasting benchmark, \textsc{TimeTextCorpus (TTC)}, achieving excellent  performance on climate and medical domains.
\end{itemize}
\section{Related Work}\label{sec:related_work}
We briefly review work related to \ourmethod{}, specifically those in multimodal time-series forecasting and Bayesian state-space models. An extended discussion on related work can be found in \autoref{appsec:extended_related_work}.

\paragraph{Multimodal Time-Series Forecasting.} Motivated by recent advancements in LLMs,  recent studies have explored their potential to enhance time-series forecasting by incorporating textual context. Some approaches use LLMs directly~\cite{merrill2024language,gruver2023large}, while others finetune them with auxiliary modules. \cite{xu2024beyond} introduce a cross-attention architecture to fuse news and time-series data. \cite{kim2024multi} propose HybridMMF, a joint model trained on the TimeTextCorpus using shared embeddings for text and time series. \cite{wang2024chattime} develop ChatTime, a foundation model fine-tuned for diverse zero-shot forecasting tasks. \cite{li2025language} present TaTS, which augments standard models with concurrent textual inputs as auxiliary variables.

\textit{These works do not explore the integration of LLMs into state space models, which is the key contribution of our work}. By incorporating LLMs into state space models, we not only improve time-series forecasting performance but also exploit the benefits of state space models, such as uncertainty estimation.

% Time-MMD, HybridMMF(TTC)

\paragraph{Bayesian State Space Models.}
Bayesian filtering has long been central to time-series modeling, with the classical Kalman filter~\cite{kalman1960new} offering optimal estimation in linear Gaussian systems. To address nonlinear and non-Gaussian dynamics, extensions such as the Extended Kalman Filter, Unscented Kalman Filter~\cite{julier2004unscented}, Ensemble Kalman Filter~\cite{houtekamer1998data}, and Sequential Monte Carlo~\cite{doucet2001sequential} have been proposed to approximate posterior distributions using linearization or sampling techniques.
More recent work combines state space structure with deep neural networks. KVAE~\cite{(kvae)fraccaro2017disentangled} and \cite{rangapuram2018deep} apply neural parameterizations to learn latent dynamics for sequential and multivariate time-series forecasting. In reinforcement learning, models like PlaNet~\cite{hafner2019learning}, Dreamer~\cite{hafner2020mastering}, and KalmanNet~\cite{revach2022kalmannet} extend these ideas to partially observable, high-dimensional control tasks.

Despite this progress, most prior work focuses on unimodal data such as images or numerical sequences. \textit{Our work is the first to integrate pretrained large language models (LLMs) with a probabilistic state space model for joint numeric and textual forecasting}, extending Bayesian state estimation into the realm of multimodal generative modeling.

%\textbf{TODO: one paragraph on classical Bayesian filters and smoothers \citep{kalman1960new}\citep{kalman1961newresults}\citep{julier2004unscented}}

%\textbf{TODO: another paragraph on machine learning papers using Bayesian estimators \citep{(kvae)fraccaro2017disentangled}\citep{hafner2019learning}\citep{hafner2020mastering}\citep{revach2022kalmannet}\citep{rangapuram2018deep}}

%Originally studied in control and decision theory, Bayesian state estimators are used to infer the true state of a dynaThe seminal work by \cite{kalman1960new}, the Kalman filter, introduced an optimal recursive algorithm for estimating unobservable states of a dynamical system from noisy measurements. While initially developed for linear systems, following works such as the Unscented Kalman Filter (UKF) applicability to 

%Assuming a mathetmatical process over time, Bayesian state estimation, also known as Bayesian filtering, is a  Widely used in real-world appications on navigation systems, robotic motion control and planning, and econometric signal processing. Our work extends the applicability of Kalman filters to textual data by incorporating an autoencoder to filter state estimates based on pseudo-observations.

\section{Preliminaries}\label{sec:background}

We provide background information on multimodal time-series forecasting (\S\ref{subsec:problem_setting}), followed by a discussion on the latent dynamical system we work with (\S\ref{subsec:state_space_models}). We also derive the multimodal extension of the evidence lower bound that we use for approximate Bayesian inference. Specific architectural choices made to realize LLM integration are discussed in the following section.

%In this section, we share the problem setting and useful notation (\S\ref{subsec:problem_setting}), followed by a brief introduction to Gaussian SSMs widely used in dynamical system identification and other unimodal TSF frameworks (\S\ref{subsec:state_space_models}). We end with an introduction on Kalman filtering and RTS smoothing that can compute closed-form posterior state estimates conditioned on the observations (\S\ref{subsec:ekf_background}).

\subsection{Multimodal Forecasting}\label{subsec:problem_setting} Given a temporal series of numerical values across $t$ time steps $\vy_{1:t} = [\vy_1, \dots, \vy_t] \in \R^{t \times M}$, the objective of \textit{unimodal} time-series forecasting  is to predict the target values of next $H$ steps $\vy_{t+1:t+H}= [\vy_{t+1}, \dots, \vy_{t+H}]$ where $H$ denotes the prediction horizon. Depending on the dimension $M$ of the prediction target, we refer to the TSF problem as univariate ($M = 1$) or multivariate ($M > 1$). In this work, we will focus on the univariate setup; we note that our method can easily be extended to multivariate scenarios.

In our \textit{multimodal} forecasting setup, we are additionally given a temporally aligned sequence of textual data $\bm{\gD}_{1:t} = [\bm{\gD}_1, \dots, \bm{\gD}_t]$. Each such point is itself a sequence of tokens $\bm{\gD}_{t} = [w_1, \dots , w_{\text{len}(\bm{\gD}_t)}]$. Using both target values and textual data, our goal is to train a mapping that predicts future target values as well as corresponding text:
\begin{align*}
    f_\Theta: (\vy_{1:t}, \bm{\gD}_{1:t}) \mapsto (\vy_{t+1:t+H}, \bm{\gD}_{t+1:t+H})
\end{align*}

%Intuitively, this multimodal formulation captures richer predictive targets, enabling both quantitative forecasts and qualitative insights. 
Note that existing multimodal forecasters are commonly trained under rigid assumptions on fixed input windows and prediction horizons with no explicit modeling of uncertainty, learning a deterministic mapping $f_{\Theta}: (\vy_{t-L:t}, \bm{\gD}_{t-L:t}) \to (\vy_{t+1:t+H}, \bm{\gD}_{t+1:t+H})$ with finite lookback size $L$ and prediction horizon $H$. This limitation reduces their adaptability to real-world scenarios.
%It is worth noting that existing Linear or Transformer-based methods typically trained to perform forecasting on a fixed horizon $H$  conditioned on a fixed lookback window of $L$ entries (\textit{i.e.}, $[\vy_{T-L+1}, \dots, \vy_{T}]$).
%Note that each document $\bm{\gD}$ consists of a sequence of tokens $\bm{\gD}_t = [w_1, \dots, w_k]$, the likelihood of which is parameterized by LLMs via next-token prediction $p(\bm{\gD_t}) = \prod_{i=1}^{\text{len}(\bm{\gD_t})} p(w_i \mid w_{<i})$.

\subsection{Bayesian State Space Models}\label{subsec:state_space_models}

%Despite the varying modality and generative processes, all observations seen at a single time point effectively originate from a shared unobservable state. Hence to make accurate forecasts, a model must be able to accurately model the underlying dynamics of the environment. To do so, we use a nonlinear Markovian SSM to model the temporal evolution of unobservable states and target observations generated from the states. 
%To overcome the aforementioned limitations, we adopt a Bayesian state-space modeling approach, which provides a principled way to capture the underlying latent dynamics governing both numeric and textual observations.
On the other hand, state space models offer a much flexible alternative to Transformer-architectures, with an inductive bias that better fits the temporal forecasting scenario. For this reason, we adopt a Bayesian state space modeling approach, providing a principled way to capture the underlying latent dynamics governing both numeric and textual observations in an uncertainty-aware manner.

\paragraph{Latent dynamical model.} At each time step $t$, we assume that a shared unobservable latent state $\vx_t \in \R^N$ encodes the system's internal condition, which evolves stochastically over time and gives rise to both types of data as conditionally independent emissions.
%To overcome these limitations, we adopt a Bayesian state-space modeling approach, which provides a principled way to capture the underlying latent dynamics governing both numeric and textual observations. At each time step $t$, we assume that a shared unobservable latent state $\vx_t \in \R^N$ encodes the system's underlying condition, which stochastically evolves over time and generates both modalities as conditionally independent emissions.

\begin{wrapfigure}{R}{0.45\textwidth}
  \vspace{-15px}
  \begin{center}
    \includegraphics[width=0.45\textwidth]{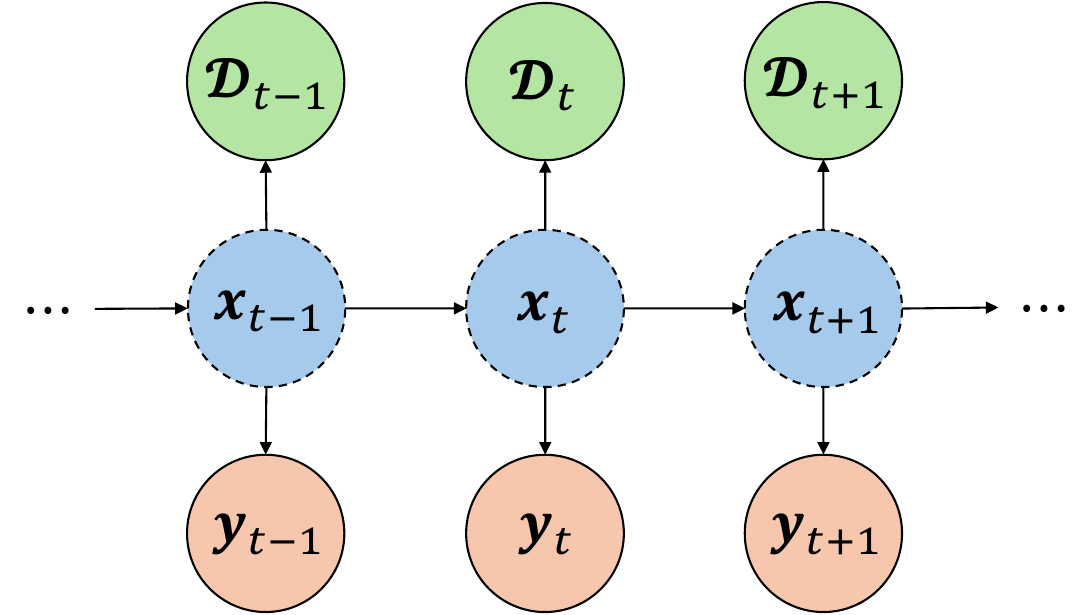}
  \end{center}
  \caption{The latent dynamical model assumed by \ourmethod{}. The SSM serves as a backbone modeling the temporal dynamics of latent states $\vx_t$, from which multimodal data $\vy_t$ and $\bm{\gD}_t$ are generated.}\label{fig:graphical_model}
  \vspace{-10px}
\end{wrapfigure}

Formally, the model is defined by three core components (\autoref{fig:graphical_model}). First, the \textbf{state transition model} $p(\vx_t \mid \vx_{t-1})$ governs the evolution of latent states over time, modeling the temporal evolution of the system. Second, the \textbf{numeric emission model} $p(\vy_t \mid \vx_t)$ captures how the target numeric observations are generated from each latent state. Third, the \textbf{textual emission model} $p(\bm{\gD}_t \mid \vx_t)$ models the generation of textual descriptions from the same latent state.

%Assuming an $N$-dimensional state space with each $\vx_t \in \R^N$ representing the underlying state at time $t$, the core components of \ourmethod{} model the following stochastic dynamics (\autoref{fig:graphical_model}):
%\begin{itemize}[leftmargin=15px]
%    \item State transition function: $\vx_{t} \sim p(\vx_{t} \mid \vx_{t-1})$
%    \item Value emission function: $\vy_{t} \sim p(\vy_t \mid \vx_{t})$
%    \item Text emission function: $\bm{\gD}_{t} \sim p(\bm{\gD}_t \mid \vx_{t})$
%\end{itemize}

For simplicity, we parameterize the transition distribution $p(\vx_t \mid \vx_{t-1})$ as a multivariate Gaussian $\mathcal{N}(\bm{\mu}_{t}, \bm{\sigma}_{t})$, where both the mean $\bm{\mu}_{t}$ and diagonal variance $\bm{\sigma}_{t}$ are produced by a recurrent neural network (e.g., GRU or LSTM) applied to the previous state $\vx_{t-1}$. The numeric observation model $p(\vy_t \mid \vx_t)$ is also modeled as a Gaussian with fixed variance, the mean of which can be computed by passing $\vx_t$ through a multi-layer perceptron (MLP). Under this formulation, maximizing the likelihood of the numeric data is equivalent to minimizing the mean squared error (MSE)~\cite{hafner2019learning}.

Modeling the likelihood of text given contextual information---textual or non-textual---is a core strength of LLMs~\cite{(llava)liu2023visual}. In our work, we leverage this capability by incorporating a pretrained LLM to model the conditional distribution $p(\bm{\gD}_t \mid \vx_t)$. By finetuning the LLM, we enable it to generate coherent and semantically meaningful text sequences $\bm{\gD}_t$ that reflect the underlying dynamics encoded in each latent state $\vx_t$. This design allows \ourmethod{} to jointly produce interpretable natural language and accurate numerical forecasts, both grounded in a shared latent trajectory.

\paragraph{Training objective.}
We train the model by maximizing the evidence lower bound (ELBO) on the joint likelihood of the observed numeric and textual data (i.e., our loss function is the negative ELBO). The classical ELBO objective naturally extends to our multimodal setting as
\begin{gather}
    \log p(\vy_{1:T}, \bm{\gD}_{1:T}) = \log \int_{\vx_{1:T}} \prod_{t=1}^T p(\vx_t \mid \vx_{t-1}) p(\vy_t \mid \vx_t) p(\bm{\gD}_t \mid \vx_t) d\vx_{1:T}\nonumber\\
    \geq \sum_{t=1}^T \E_{q(\vx_t \mid \vy_t,\bm{\gD}_t)} \bigg[\underbrace{\log p(\vy_t \mid \vx_t)}_{\text{value likelihood}} + \underbrace{\log p(\bm{\gD}_t \mid \vx_t)}_{\text{text likelihood}}\bigg] - \underbrace{\text{KL}(q(\vx_t \mid \vy_{1:t},\bm{\gD}_{1:t}) \mid\mid p(\vx_t \mid \vx_{t-1}))}_{\text{temporal regularization}} \label{eqn:objective}
\end{gather}
Replacing the computationally intractable posterior distribution $p(\vx_t \mid \vy_{1:t}, \bm{\gD}_{1:t})$, we introduce a variational posterior $q(\vx_t \mid \vy_{1:t}, \bm{\gD}_{1:t})$ over the latent states, parameterized via a deep Kalman filter~\cite{krishnan2015deep,haarnoja2016backprop,hafner2019learning}.
%This posterior inference network conditions on the entire history of observations up to time $t$, leveraging both numeric and textual cues to produce temporally coherent state estimates. The training objective effectively balances accurate reconstruction of both modalities and adherence to prior dynamics defined by the transition model. 
Intuitively, the training objective effectively balances three essential aspects of Bayesian state estimation. First, the expected likelihood terms ensure fidelity to the observed data by encouraging the latent states to retain enough information to accurately reconstruct both the numeric values and textual descriptions. Second, the KL regularizer imposes temporal coherence by penalizing latent trajectories that deviate too strongly from the prior dynamics controlled by the SSM. Lastly, the variational expectation allows the model to predict under uncertainty in the latent trajectory, inducing more robust and generalizable forecasts. 
%Algorithm~\ref{alg:training_step} shares the training procedure in pseudocode, and 
A full derivation of \autoref{eqn:objective} can be found in the appendix.

\begin{minipage}[t]{.56\linewidth}
    \vspace*{0mm}   
    \IncMargin{1.5em}
\begin{algorithm}[H]
  \SetAlgoLined
  \SetKwInOut{Input}{Input}
  \SetKwInOut{Output}{Output}
  \Indm 
  \Input{ Previous state and hidden $(\hat{\vx}_{t-1}, \vh_{t-1})$ \\\vspace{0.2em} \,Current text and value $(\bm{\gD}_{t}, \vy_{t})$}
  \vspace{0.3em}
  \Output{ Current state and hidden $(\hat{\vx}_{t}, \vh_{t})$}
  \vspace{0.3em}
  \Indp 
  Get prior $\mathcal{N}(\bm{\mu}_{t},\bm{\sigma}_{t}), \vh_{t} = \text{SSM}(\hat{\vx}_{t-1}, \vh_{t-1})$\\
  \vspace{0.3em}
  Summarize text $\vs_{t} = \text{LLM}_{\text{encoder}}(\bm{\gD}_{t})$\\
  \vspace{0.3em}
  Get posterior $\mathcal{N}(\hat{\bm{\mu}}_{t},\hat{\bm{\sigma}}_{t}) = \text{MLP}_{\text{post}}(\vh_{t}, \vy_{t}, \vs_{t})$\\
  \vspace{0.3em}
  Sample $\hat{\vx}_{t} \sim \mathcal{N}(\hat{\bm{\mu}}_{t}, \hat{\bm{\sigma}}_{t})$ via reparameterization\\
  \vspace{0.3em}
  $\gL_{\text{val}} = \|\vy_{t} - \text{MLP}_{\text{val}}(\hat{\vx}_{t})\|^2$\\
  \vspace{0.3em}
  $\gL_{\text{text}} = \text{LLM}_{\text{decoder}}(\hat{\vx}_{t}, \bm{\gD}_{t})$\\
  \vspace{0.3em}
  $\gL_{\text{KL}} = \text{KL}(\mathcal{N}(\hat{\bm{\mu}}_{t},\hat{\bm{\sigma}}_{t}) \mid\mid \mathcal{N}(\bm{\mu}_{t},\bm{\sigma}_{t}))$\\
  \vspace{0.3em}
  Update parameters via $\gL = \gL_{\text{val}} + \gL_{\text{text}} + \gL_{\text{KL}}$\\
  \vspace{0.3em}
  \Return $(\hat{\vx}_{t}, \vh_{t})$
  \caption{Stateful training step of \ourmethod{}}
  \label{alg:training_step}
\end{algorithm}
\DecMargin{1.5em}
\end{minipage}
\hfill
\begin{minipage}[t]{.40\linewidth}
    \vspace*{0mm}
    %\vspace{-5px}
    \includegraphics[width=\linewidth]{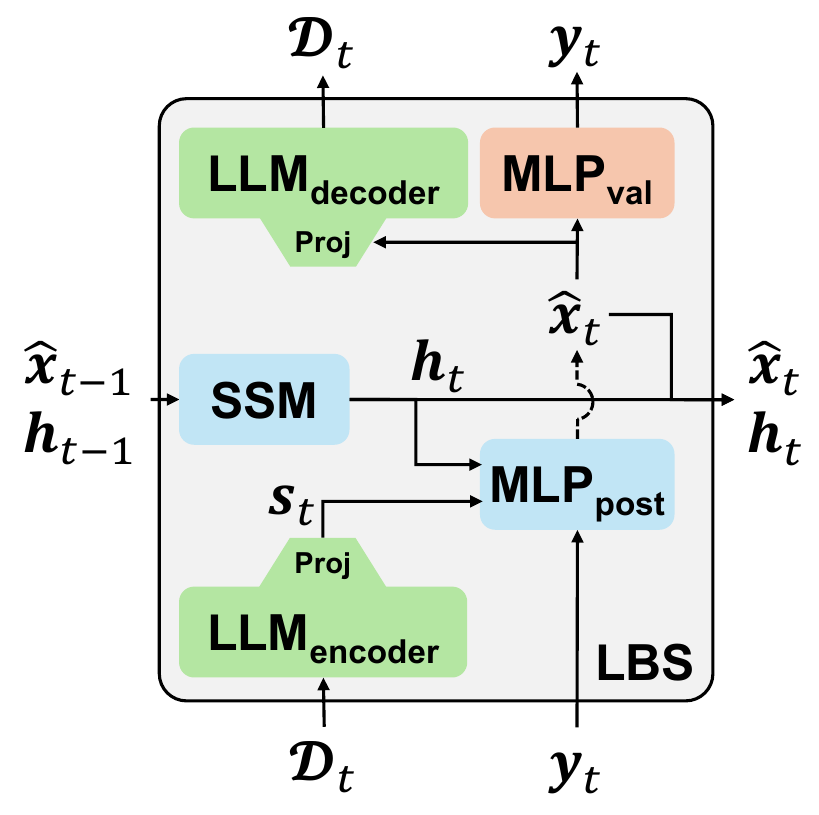}
    \vspace{-13px}
    \captionof{figure}{Illustration of a single forward pass through \ourmethod{}.}\label{fig:singlestep_pipeline}
\end{minipage}

\section{\ourmethod{}: LLM-Integrated Bayesian State Space Model}\label{sec:method}

%For climate forecasting, weather patterns at a certain area are determined by exact rules of geophysical fluid dynamics, which operate on a state space too large and are computationally expensive to be simulated exactly. Nonetheless, the states at each time point can be recognized indirectly under observational noise via atmospheric measurements (\textit{e.g.}, temperature, humidity, etc.) and weather reports in text. Despite the varying modality and generative processes, all observations seen at a single time point effectively originate from a shared unobservable state.

To integrate textual information into our latent dynamical model, we must address two technical challenges: \textbf{(C1) How can we design the LLM-based filter that estimates the a posteriori state conditioned on text (i.e., $q(\vx_t \mid \vy_t, \bm{\gD}_t)$)?}  \textbf{(C2) How can we model the likelihood of text conditioned on the latent state (i.e. $p(\bm{\gD}_t \mid \vx_t)$)?} In this section, we detail the architectural components that tackle these two challenges. The components form the foundation of our proposed framework \ourmethod{}. An illustration of the architecture can be found in \autoref{fig:autoencoder}.
%To optimize our latent dynamical model under textual observations, we need to solve two methodological questions: \textbf{Q1: how can we perform posterior state estimation conditioned on textual data using an LLM?} (i.e., how to model $q(\vx_t \mid \vy_t, \bm{\gD}_t)$?) and \textbf{Q2: how can we generate text given the unobserved state?} (i.e., how to model $p(\bm{\gD}_t \mid \vx_t)$?) Here we present the architectural designs made to enable both directions, together leading to our \ourmethod{} framework.

%Based on this notion, \ourmethod{} trains a Gaussian state-space model to represent the underlying dynamics and serve as the temporal backbone from which both numeric and textual observations are realized. In the following sections, we provide details on the LLM-Grounded Text Autoencoder that enables closed-form Bayesian filtering (\S\ref{subsec:autoencoder}). We then present different components of the multimodal GSSM assumed in \ourmethod{}(\S\ref{subsec:state_space_model}), followed by details on how \ourmethod{} obtains a posteriori state estimates and how GSSM parameters are tuned (\S\ref{subsec:training}).

\subsection{Text-conditioned Posterior State Estimation}\label{subsec:encoder}

%An immediate question that arises when filtering with textual data is how to fit non-numerical data such as text for posterior estimation. 
%To estimate the a posteriori state conditioned on the text, one possible method would be to optimize the single-step Gaussian prior and by maximizing the evidence lower-bound~\citep{(vae)kingma2013auto}. However, this would imply that each filtering time-step requires gradient-based optimization via multiple forward-passes through the LLM, which would be computationally intractable when modeling a sequence of textual data.

%One approach to estimate the posterior distribution of latent states given textual input would involve optimizing a timestep-specific ELBO using gradient-based inference, where the latent state is inferred by maximizing the ELBO conditioned on the text observation~\cite{(vae)kingma2013auto}. However, this requires iterative optimization involving multiple forward passes through a large language model (LLM) per time step, which is computationally prohibitive for long sequences or real-time applications. 
To efficiently update our prior state estimates conditioned on text, we draw inspiration from recent work~\cite{(autocompressor)chevalier2023adapting,(icae)ge2023context} and adapt the pretrained LLM for  context compression towards fast one-shot encoding of textual observations into latent state summaries. 
Then, we forward latent state summaries together with the prior state $\vx_t$ and numeric value $\vy_t$ for posterior inference.

\paragraph{Text Compression.} The core idea is to introduce special tokens unique to the task of summarization, and finetune the LLM to store critical information in the summary tokens for effective posterior inference. Concretely, we first augment the vocabulary of the pretrained LLM with $K$ special learnable tokens $\texttt{<SUM>}_k$, which facilitate the task of text compression. To encode a textual observation $\bm{\mathcal{D}}_t$, we append all $K$ summary tokens after the input sequence $\bm{\mathcal{D}}_t$, and forward the augmented sequence through the pretrained LLM. While also possible to randomize the positions of summary tokens during training to mitigate positional biases~\cite{(autocompressor)chevalier2023adapting}, we empirically found that it leads to negligible changes in performance, possibly due to lack of variance in text length.
%To leverage pretrained generative capabilities of LLMs for effective representation learning, we incorporate parts of the pretrained LLM throughout the autoencoding process. 

After processing the sequence through the LLM, we extract the final hidden states of the $K$ inserted summary tokens, then concatenate along the feature dimension to form a single summary vector of $\bm{\gD}_t$. This vector is then projected through a MLP to obtain a low-dimensional summary vector $\vs_t \in \mathbb{R}^{N}$ that matches the latent states in dimension.

\paragraph{Posterior Inference.} Given the summary vector $\vs_t$, we compute the mean and diagonal covariance of the variational posterior distribution $q(\vx_t \mid \vy_t, \vs_t)$, assumed to be Gaussian, via a neural Kalman filter parameterized by another MLP~\cite{krishnan2015deep}. This MLP takes as input the summary vector $\vs_t$, the corresponding numeric target $\vy_t$, the prior latent state $\vx_{t}$, and outputs the mean and log-variance of the posterior distribution. Note the entire process is end-to-end trainable, hence we finetune the LLM using LoRA~\citep{hu2022lora} to effectively encode forecasting-relevant semantic and temporal information into the $\texttt{<SUM>}_k$ tokens without significantly altering the generative capabilities within its pretrained weights. The detailed prompt used to compress text can be fo und in the appendix.

\subsection{State-conditioned Text Generation}

%LLaVA visual instruction tuning. Global prefix reinforces the representational bottleneck. It is shared across all time steps, so global information about the climate/medical must be encoded into these embeddings, the more time-step information must be inferred from the states. Difference is can the model and state tune itself to draw a temporally consistent sequence of textual data?

Given the posterior distribution, we use the reparameterization trick~\cite{(vae)kingma2013auto} to generate $\hat{\vx}_t \sim q(\vx_t \mid \vy_t, \bm{\gD}_t)$, generating Monte-Carlo samples in an end-to-end learnable manner. 
Then, we can model the posterior state-conditioned textual likelihood $p(\bm{\gD}_t \mid \hat{\vx}_t)$ by providing a projection of $\hat{\vx}_t$ as a prefix to the LLM~\cite{li2021prefix}, similarly to vision-text instruction tuning frameworks~\cite{(llava)liu2023visual}. 
%Then the posterior state-conditioned textual likelihood $p(\bm{\gD}_t \mid \hat{\vx}_t)$ can be modeled by providing a projection of $\hat{\vx}_t$ as a prefix. 
%Generating text grounded on non-textual contexts has been an active area of research: for instance, models like LLaVA~\citep{(llava)liu2023visual} achieve visual instruction following by encoding image features and feeding them as soft prompts (i.e., token embeddings) into a pretrained LLM. 
%This approach allows the LLM to condition generation on rich visual context while preserving its language modeling capabilities.
%Inspired by such strategy, we enable state-conditioned text generation by treating the latent states produced by the probabilistic SSM as conditioning context for the LLM. 
Specifically, we project the sampled low-dimensional latent state vector $\hat{\vx}_t$ into a sequence of tokens for the LLM using a linear layer. These projected tokens are then prepended to $\bm{\gD}_t$, effectively allowing the LLM to condition its generation on the state dynamics captured by the temporal SSM backbone.

%However, because the latent states $\vx_t$ reside in a low-dimensional space, a simple projection may be insufficient to capture the full range of semantic and textual variations present in data. To mitigate this limitation, we introduce a set of global instruction tokens shared across all time steps. Ideally, these learnable embeddings serve as a soft instruction prompt that guides the LLM to interpret the projected state embeddings in a forecast-oriented manner and to learn global textual patterns and formatting conventions consistent across the sequence~\cite{li2021prefix}. 

This design assumes that the temporal backbone is capable of encoding time-specific information such as event structure, trends, or contextual shifts within a compact latent space~\cite{cai2024timeseriesexam}. By projecting this information into the high-dimensional language space and augmenting it with instruction prompts, we provide the LLM with the necessary information to generate fluent and temporally consistent text.

\paragraph{Shared or Separate LLMs?} While it is possible to use two separate LLMs for encoding and decoding, for \ourmethod{} we assume the same LLM weights are shared between the two steps: the LLM that encodes text into compressed embeddings for posterior estimation also serves as the decoder for text generation. This weight sharing not only reduces the computational burden, but also encourages the LLM to encode forecasting-relevant information in a way that it can later reuse for generation, learning prediction and inference in a self-consistent manner.
%Note that \ourmethod{} while the LLM that encodes textual inputs into summary embeddings for posterior state estimation also serves as the decoder for text generation. This weight sharing not only reduces the computational burden, but also encourages the LLM to encode forecasting-relevant information in a manner that itself can later reuse for generation, enabling inference and prediction in a self-consistent manner.

%\subsection{Implementation Details \todo{will move to appendix}}\label{subsec:training}

%While integrating an LLM with SSM enables new capabilities, it also presents unique challenges in implementation. Therefore we close the section by discussing details of the training pipeline that enable tractable and effective training of \ourmethod{}.

\paragraph{Stateful single-step training.} Ideally, training \ourmethod{} on long sequences would allow the model to better capture long-range dependencies. However, each timestep $t$ requires passing $\bm{\gD}_t$ through the LLM twice---once for encoding and once for decoding---making na\"ive long-horizon training computationally expensive. To address this, we adopt \textit{stateful} training~\cite{yen2022stateful,katrompas2022enhancing}, where the model is trained on single-step batches under its temporal ordering, with hidden states passed onto the next training iteration. The detailed algorithm and illustration of a single training step can be found in Algorithm~\ref{alg:training_step} and \autoref{fig:singlestep_pipeline}.

\section{Experimental Results}\label{sec:experiments}

%\subsection{Synthetic Textual Data}

We empirically explore the following claims about \ourmethod{}.
\begin{itemize}[leftmargin=15px]
    \item \textbf{Claim 1: \ourmethod{} improves performance in multimodal time-series forecasting. (\S5.1)}
    \item \textbf{Claim 2: \ourmethod{} can model predictive uncertainty using its probabilistic architecture. (\S5.2)}
    \item \textbf{Claim 3: \ourmethod{} encodes trends and seasonality within its latent states. (\S5.3)}
    \item \textbf{Claim 4: Scaling the LLM in \ourmethod{} improves performance. (\S5.4)}
\end{itemize}

\begin{figure}[!t]
    \centering
    \includegraphics[width=\linewidth]{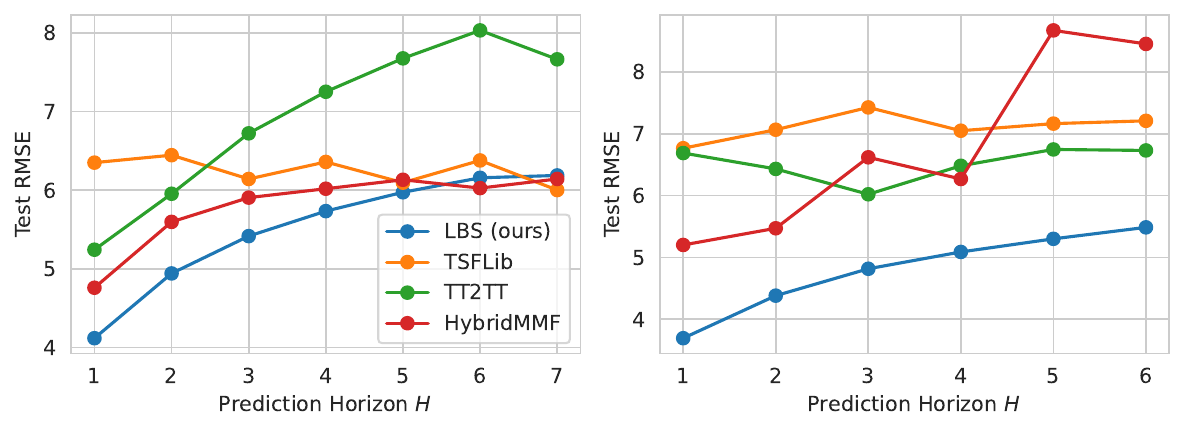}
    \caption{Test RMSE results from \textsc{TTC-Climate} (left) and \textsc{TTC-Medical} (right). }
    \label{fig:ttc_main_results}
\end{figure}

\paragraph{Datasets.} We perform experiments on the \textsc{TextTimeCorpus} (\textsc{TTC}~\cite{kim2024multi}), a multimodal time-series forecasting benchmark that covers two distinct domains: \textsc{TTC-Climate} consists of daily temperature measurements at Washington DC with textual weather descriptions. \textsc{TTC-Medical} consists of daily heart rate measurements from hospitalized patients accompanied by nursing notes. Following previous work~\cite{kim2024multi}, we use a 8-1-1 train-validation-test split across time. Further details on the datasets can be found in the appendix.

%\paragraph{Datasets.} We perform experiments on the \textsc{TextTimeCorpus} (\textsc{TTC}~\cite{kim2024multi}), a multimodal time-series forecasting benchmark that covers two distinct domains: \textsc{TTC-Climate} is consisted of daily temperature measurements at Washington DC, accompanied by textual weather descriptions spanning from January 1st, 2014 to December 1st, 2023. \textsc{TTC-Medical} stores daily heart rate measurements from 73 patients accompanied by nursing notes writing observations and treatment plans. Each patient data spans an average length of 104 days. Example texts from each dataset can be found in the appendix. Following previous work~\cite{kim2024multi}, we train the model on the first 80\% of all timestamps, validate on the next 10\%, then test on the last 10\%. 

%\input{tables/ttc_climate_main_results}
%\input{tables/ttc_medical_main_results}
%\input{tables/ttc_combined_main_results}
%\input{tables/ttc_climate_text_results}

\subsection{Time-Series Forecasting Performance}

%We first evaluate the quantitative forecasting performance of \ourmethod{} against existing baselines.

\paragraph{Setup.} We first compare our model against existing multimodal forecasters TSFLib~\cite{liu2024time}, TextTime2TextTime (TT2TT~\cite{kim2024multi}), and HybridMMF~\cite{kim2024multi}. For TSFLib, we use Reformer as its time-series forecasting backbone, as it was the best-performing setup. We also compare against unimodal methods, results from which we defer to \autoref{appsec:extended_results} for brevity. Note that all baselines except \ourmethod{} are specifically trained for each prediction horizon $H$. On the other hand, a single \ourmethod{} model is evaluated on all possible $H$ to demonstrate its generalizability to arbitrary forecast horizons. All multimodal models adopt LLaMA3.1-8B~\cite{grattafiori2024llama} as the base LLM, and \ourmethod{} uses a single-layer GRU~\cite{(gru)cho2014learning} with latent dimension 16 as the SSM backbone. Further details on training and model hyperparameters can be found in the appendix.

%\paragraph{Expected Results.} As SSMs enjoy an inductive bias that better fits temporal forecasting problems, we hypothesize \ourmethod{} to outperform non-SSM based models even in unimodal scenarios. By additionally incorporating textual data, we expect \ourmethod{} to further improve in performance throughout all horizons, given that the information within text are conducive to accurate forecasting.
%\begin{table}[!t]
\begin{wraptable}{R}{.5\textwidth}
    \vspace{-12px}
    \centering
    \caption{Test RMSE and relative improvements from incorporating textual data in \ourmethod{}, averaged across all prediction horizons.}\label{tab:unimodal_vs_multimodal}
    \resizebox{.5\textwidth}{!}
    {
    \begin{tabular}{c|cc}
        \toprule
        Method & \textsc{TTC-Climate} & \textsc{TTC-Medical}\\
        \midrule
        Unimodal & 5.644 & 4.992 \\
        Multimodal & 5.422 & 4.733\\
        \midrule
        $\Delta$ (\%) & 3.815 & 5.410\\
        \bottomrule
    \end{tabular}
    }
\end{wraptable}   
%\end{table}
\paragraph{Results.} \autoref{fig:ttc_main_results} presents forecasting results across varying prediction horizons. \textbf{For most prediction horizons, \ourmethod{} achieves substantial gains over all baselines}, improving the best performances by 5.13\% and 21.28\% on \textsc{TTC-Climate} and \textsc{TTC-Medical} on average, respectively. Combined with the fact that a single \ourmethod{} model is evaluated throughout all horizons, this result highlights the strength and generalizability of SSMs in capturing temporal dependencies, validating our choice of using a probabilistic SSM as our temporal backbone. When compared with a unimodal variant of \ourmethod{} that does not use textual data, \autoref{tab:unimodal_vs_multimodal} shows that multimodal \ourmethod{} enjoys improvements in performance overall due to the additional modality. 
%, which offers a strong inductive bias for structured temporal sequence modeling. 
%\textbf{When compared with unimodal \ourmethod{}, multimodal \ourmethod{} enjoys further improvements in predictive performance across all horizons by incorporating textual data}, leading to an average RMSE reduction of 0.141 on \textsc{TTC-Climate}, 0.197 on \textsc{TTC-Medical}. 
This highlights the model’s ability to leverage textual information for more accurate posterior inference, leading to sharper and more informed forecasts.
%This demonstrates its ability to leverage textual data to infer more accurate posterior estimates, which in turn enables sharper and more informed future predictions. 
%The consistent improvements across both domains emphasize that textual information, when properly fused via the latent states, carries significant complementary signal for downstream forecasting.

\paragraph{Text Generation.} Beyond forecasting numeric values, \textbf{\ourmethod{} is also capable of generating temporally coherent textual forecasts}. \autoref{fig:climate_text_example} shows a sample forecast generated by \ourmethod{} on the \textsc{TTC-Climate} dataset. While only provided with the prior state embedding and a simple prompt, \ourmethod{} can produce context-aware descriptions that align well with the ground-truth dynamics without direct access to previous text. This result highlights the latent states' ability to encode rich semantic structure that can further rationalize model forecasts, and also demonstrates its utility in applications where human-readable justifications are essential alongside quantitative predictions.

\begin{figure}[!t]
    \centering
    \begin{tcolorbox}[title={\footnotesize Ground Truth},boxsep=1pt,left=2pt,right=2pt,top=1pt,bottom=1pt]
    \scriptsize \hlc[red!30]{Cooler than normal temperatures in the central and eastern U.S., with warmth in the western U.S.} A couple of rounds of \hlc[orange!30]{heavy rain expected in the East, particularly from Friday into Saturday}, with moisture from the Gulf of Mexico and western Atlantic contributing to rainfall. Heavy rainfall risks are present for the Mid-Atlantic and southern Plains, with potential local runoff issues due to expected convection. A significant closed upper low/trough will bring much below normal temperatures and an unsettled pattern across the north-central U.S. and Great Lakes. Much \hlc[green!25]{above average temperatures with some record values possible in the West Coast and Interior West}. High winds anticipated in the Central Rockies, Central Plains, and Northern Rockies on Friday, April 28. \hlc[cyan!30]{Flooding concerns exist in the Upper and Middle Mississippi Valleys}, along with parts of the Northern Plains and Great Basin.
    \end{tcolorbox}
    \vspace{-5px}
    \begin{tcolorbox}[title={\footnotesize Output from \ourmethod{}},boxsep=1pt,left=2pt,right=2pt,top=1pt,bottom=1pt]
    \scriptsize From April 28 to May 2, 2023, expect a persistent weather pattern with \hlc[red!30]{troughing in the eastern U.S. and ridging in the western U.S}. This will lead to \hlc[orange!30]{a wet pattern in the east} and dry conditions in the west. A deep cyclone will track from the Midwest to the Northeast, causing moderate to heavy rainfall, particularly in \hlc[cyan!30]{the Ohio Valley, Appalachians, and Northeast, with potential flooding}. Light to moderate precipitation is expected across the Midwest, Great Lakes, and Northeast, with \hlc[orange!30]{the heaviest rainfall on Friday and Saturday}. In contrast, \hlc[green!25]{the western U.S. will experience well above normal temperatures} (10-20°F above normal) with little to no precipitation, increasing drought concerns. A cold front will bring cooler temperatures to the east, with the first half of May likely seeing above normal temperatures. Flooding is possible in the Southern Plains and the Pacific Northwest, with severe weather forecasted for the Southern Plains on April 28.
    \end{tcolorbox}
    \vspace{-5px}
    \caption{Example text comparison generated by \ourmethod{} vs. ground truth text from \textsc{TTC-Climate}. \ourmethod{} is able to textually forecast key characteristics by contextualizing the LLM on the latent states.}
    \label{fig:climate_text_example}
\end{figure}

\begin{wrapfigure}{R}{0.45\textwidth}
  \vspace{-37px}
  \begin{center}
  \includegraphics[width=0.44\textwidth]{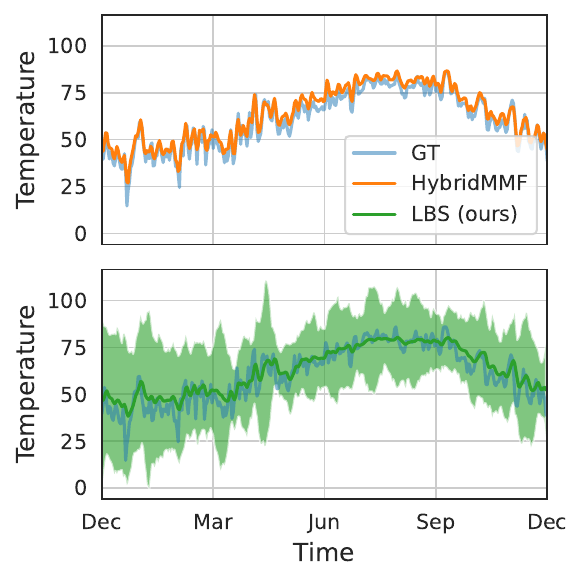}
  \end{center}
  \vspace{-10px}
  \caption{Single-step predictions ($H = 1$) of HybridMMF (top) and \ourmethod{} (bottom) on the \textsc{TTC-Climate} test set. The shaded region indicates the variance of each prediction of \ourmethod{}, with true values shown in light blue. Forecasts in the initial winter exhibits relatively larger variance than in the summer, as expected from the high variance in actual data.}\label{fig:uncertainty_plots}
  \vspace{-40px}
\end{wrapfigure}
\begin{figure}[!t]
    \centering
    \includegraphics[trim={7px 5px 5px 5px},clip,width=\linewidth]{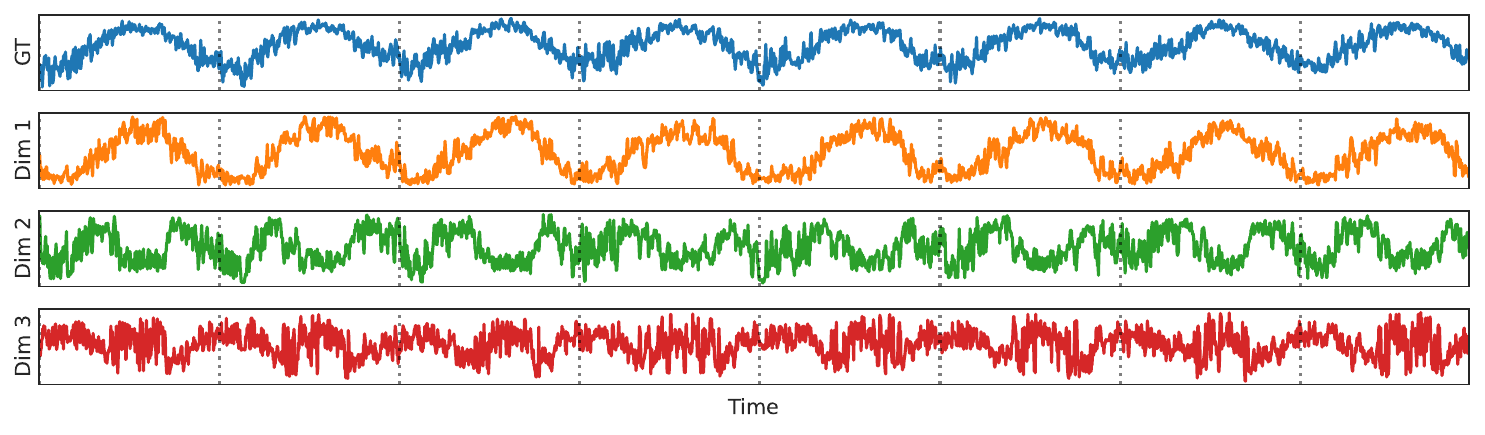}
    \vspace{-15px}
    \caption{Visualization of ground-truth signals and three t-SNE components of the state trajectory during training on \textsc{TTC-Climate}. The dashed lines indicate yearly intervals. \ourmethod{} learns states that exhibit the same seasonal patterns as the target values, promoting model transparency.}
    \label{fig:ttc_latent_trends}
    \vspace{-10px}
\end{figure}

\subsection{Uncertainty in Forecasts}

\paragraph{Setup.} In order to observe how \ourmethod{} allocates uncertainty across forecasting, we compute and report the variance in predictions across 10 states sampled from the prior distribution at each step, during test time on \text{TTC-Climate}. We compare results from \ourmethod{} against those from deterministic HybridMMF.

%\paragraph{Expected Results.} As \ourmethod{} is built on a probabilistic state-space model, it is capable of distinguishing systematic patterns against stochastic noise. As such, we hypothesize \ourmethod{} to express low uncertainty in predictable regions with low variation, higher uncertainty in regions with higher variation.

\paragraph{Results.} \autoref{fig:uncertainty_plots} shows that in contrast to deterministic baselines such as HybridMMF, \textbf{\ourmethod{} provides meaningful uncertainty intervals in addition to accurately capturing the overall trend}. We observe that the predicted variance increases in regions where the ground-truth data shows higher fluctuation (e.g., the early winter), while periods with lower fluctuation leads to lower predicted variance (e.g., the summer). This property makes \ourmethod{} particularly suitable for real-world forecasting tasks that require assessing confidence in predictions for risk-aware decision making.

\subsection{Analysis on Latent State Trajectory}

\paragraph{Setup.} To evaluate the qualitative dynamics of states learned by \ourmethod{}, we extract the posterior latent state trajectory learned by \ourmethod{} on the training set of \textsc{TTC-Climate}. For visualization, we apply t-SNE~\cite{(tsne)van2008visualizing} to the full trajectory and plot the top three components with the highest variance.
%\paragraph{Setup.} Once we train \ourmethod{} with LLaMA3.1-8B on \textsc{TTC-Climate}, we gather the posterior state trajectory from the SSM on the training data, then visualize top-$k$ components obtained decompositing the states via t-SNE.

%\paragraph{Expected Results.} We hypothesize that the low-dimensional latent trajectories reflect the same temporal characteristics of original data such as long-term trends and seasonal patterns.

\paragraph{Results.} As shown in \autoref{fig:ttc_latent_trends}, \textbf{the latent states in \ourmethod{} exhibit strong seasonal periodicity that is closely aligned with the ground-truth signal}. 
This alignment promotes transparency: the learned states are not black-box embeddings but instead encode temporally coherent structure and semantics. Such feature supports straightforward validation of the learned dynamics and enables effective diagnosis of potential errors, especially useful in high-stakes scenarios such as finance or healthcare.

%\input{tables/ttc_combined_main_results}

%\paragraph{How is uncertainty in text reflected during inference?}

%\paragraph{How well does the model perform under missing data?}

\subsection{Effect of LLM Scaling}

%As larger LLMs are known to more effectively compress information into compact summaries~\citep{(autocompressor)chevalier2023adapting, (icae)ge2023context}, we verify whether increasing the LLM size also improves forecasting performance.

\paragraph{Setup.} As larger LLMs are known to more effectively compress information into compact summaries~\citep{(autocompressor)chevalier2023adapting, (icae)ge2023context}, we verify whether increasing the LLM size also improves forecasting performance by evaluating \ourmethod{} using a range of backbone LLMs with varying parameter sizes. While larger LLMs are known to exhibit stronger reasoning and generation capabilities, it remains unclear whether these benefits translate to the setting of text-conditioned time-series forecasting. We fix our evaluation domain to \textsc{TTC-Climate} and train \ourmethod{} while switching the LLM within variants of LLaMA3 (1B, 3B, 8B)~\cite{grattafiori2024llama} and Qwen2.5 (0.5B, 1.5B, 3B, 7B)~\cite{yang2024qwen2}.

%\paragraph{Expected Results.} As larger LLMs are known to more efficiently and effectively compress information into compact summaries~\citep{(autocompressor)chevalier2023adapting, (icae)ge2023context}, we hypothesize that increasing model size should improve forecasting performance---particularly since the number of summary tokens is fixed and thus compression efficiency matters.

\begin{figure}[!t]
    \centering
    \includegraphics[width=\linewidth]{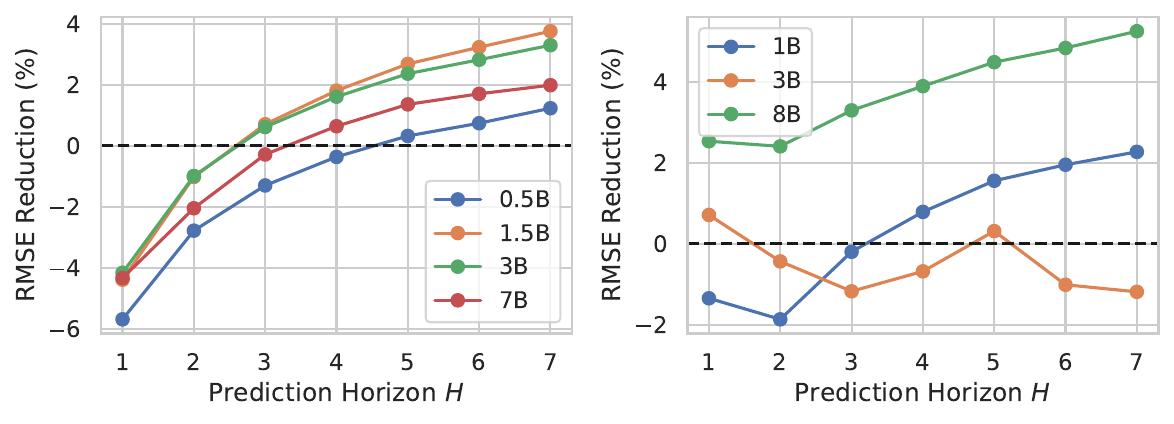}
    \caption{Test RMSE reductions of \ourmethod{} relative to its unimodal counterpart on \textsc{TTC-Climate} with varying LLMs from the Qwen2.5 (left) and LLaMA3 (right) series. The dashed line indicates the baseline from unimodal \ourmethod{}. A larger LLM does not consistently lead to better performance, but the gain from textual inputs tends to increase with increasing prediction horizon.}
    \label{fig:llm_ablation}
\end{figure}

\paragraph{Results.} Surprisingly, \autoref{fig:llm_ablation} shows that \textbf{scaling the LLM does not necessarily lead to better forecasting performance}: for instance, Qwen2.5-7B is consistently outperformed by its 1.5B variant. There are several plausible explanations. First, the relatively low capacity of the SSM may introduce a representational bottleneck, preventing \ourmethod{} from fully leveraging the richer representations offered by larger LLMs. Second, the task of compressing text into a single significantly lower dimensional vector followed by textual forecasting may not benefit from scaling as with more conventional language tasks such as question answering or code generation~\citep{roberts2025compute}. Finally, larger LLMs may tend to memorize training patterns rather than learn generalizable forecasting strategies, diminishing the role of the dynamical model.

%Surprisingly, our results reveal no consistent trend of performance improvement with larger LLMs. For instance, Qwen2.5-7B being outperformed by its 1.5B variant. We believe there are several contributing factors. First, the relatively small size of the state-space model may act as a representational bottleneck, limiting the expressiveness of the latent dynamics regardless of LLM capacity. Second, tasks involving text summarization into state vectors and subsequent text generation may not follow the same scaling behavior as traditional language tasks like QA or code generation, which typically benefit more directly from increased LLM size~\citep{roberts2025compute}. Third, larger LLMs may overfit to the training corpus by encoding memorized patterns within their parameters, reducing the burden on the SSM to learn generalizable dynamics.

Nonetheless, we observe an encouraging overall trend: \textbf{the performance gain from incorporating textual information tends to grow with longer prediction horizons}. This suggests that textual information offers complementary context that helps stabilize forecasts over time, making them more robust to compounding noise in autoregressive dynamics.

In summary, our findings highlight potential directions to better integrate LLMs for multimodal time-series forecasting: better posterior estimation strategies or capacity-aligned training of SSMs could allow larger LLMs to be used more effectively.

%These findings suggest that simply scaling the language model does not guarantee better performance in multimodal forecasting tasks and point to important future directions. For example, better integration strategies, state-aware LLM training, or capacity-aligned SSMs could allow larger LLMs to be used more effectively in this setting.

%However, we see an overall trend where the gain from incorporating textual data increases as we increase the prediction horizon, indicating that the added information from textual data can lead to predictions more robust under noise being added at each step.

%\input{figures/ablation_dimensions}
\section{Concluding Remarks}\label{sec:conclusion}

We propose \ourmethod{}, a novel architecture that integrates a Bayesian SSM with pretrained LLMs for multimodal time-series forecasting. By grounding both numeric and textual observations in a shared latent dynamical system, \ourmethod{} enables coherent forecasting along with uncertainty estimation and flexible prediction horizons. Experiments on the \textsc{TTC} benchmark demonstrate that \ourmethod{} outperforms existing baselines, with textual data providing greater gains at longer forecasting horizons. Our findings highlight the promise of probabilistic, LLM-integrated SSMs for robust and interpretable forecasting in real-world scenarios.

\paragraph{Limitations.} While \ourmethod{} generalizes well to long-horizon forecasts despite being trained with single-step batches, performance may further improve with multi-step training strategies such as backpropagation through time~\cite{werbos1990backpropagation} or latent overshooting~\cite{hafner2019learning}. However, these techniques currently pose computational and compatibility challenges with LLM optimization frameworks like DeepSpeed~\cite{rasley2020deepspeed}. Additionally, we leave integration with newer state space architectures such as Mamba~\cite{gu2023mamba} to future work, as optimized implementations for stateful training are still in development. We view these as promising directions to further enhance the scalability and flexibility of our framework.

%%%%%%%%%%%%%%%%%%%%%%%%%%%%%%%%%%%%%%%%%%%%%%%%%%%%%%%%%%%%
%\newpage
%\appendix
%\section{Technical Appendices and Supplementary Material}
%Technical appendices with additional results, figures, graphs and proofs may be submitted with the paper submission before the full submission deadline (see above), or as a separate PDF in the ZIP file below before the supplementary material deadline. There is no page limit for the technical appendices.

%%%%%%%%%%%%%%%%%%%%%%%%%%%%%%%%%%%%%%%%%%%%%%%%%%%%%%%%%%%%

%\newpage
\clearpage
\section*{Broader Impact Statement}
While our model demonstrates the potential of integrating LLMs with state space models for multimodal forecasting, it also raises important societal considerations. In particular, the ability of LLMs to generate coherent, plausible-sounding text conditioned on structured signals---such as forecasts or measurements---poses a risk of misuse. For instance, LLM-generated textual forecasts could be misinterpreted as authoritative or factual, even in the absence of uncertainty quantification or proper context. In high-stakes domains like healthcare or climate science, this may contribute to the spread of misinformation if such outputs are shared uncritically. To mitigate this, it is essential to approach text outputs with caution, clearly communicate model limitations, and consider mechanisms for human oversight and fact verification. As models like ours grow in power and accessibility, fostering responsible use and transparent reporting practices will be critical to ensure positive societal impact.

\bibliographystyle{abbrv}
\bibliography{neurips_2025}

\newpage
\appendix
\section{Extended Related Work}\label{appsec:extended_related_work}
\paragraph{Multimodal Time-Series Forecasting.} With the recent advancements in LLMs, several approaches have emerged to integrate language models with time-series forecasting. \citep{jia2024gpt4mts} introduces a textual data collection pipeline and a modified transformer architecture that uses pre-trained transformers. \cite{merrill2024language} examines whether LLMs can perform zero-shot time series forecasting with the aid of textual data, extending \cite{gruver2023large}, and concludes that even frontier models still perform poorly. \cite{xu2024beyond} proposes the Text-Guided Time Series Forecasting framework, which integrates news and descriptive textual data for time-series forecasting and introduces a new architecture that leverages a cross-attention layer for modality fusion. \cite{liu2024time} presents the Time-MMD benchmark for evaluating text-time series multimodal models and demonstrates that incorporating additional textual data can improve time-series forecasting. \cite{wang2024news} develops a reasoning agent for selecting and analyzing textual (news) data, streamlining the text processing pipeline for multimodal time-series forecasting. \cite{kim2024multi} develops the TimeText Corpus (TTC), a time-aligned text and time-series dataset for multimodal forecasting, along with a hybrid forecasting model (HybridMMF) that jointly predicts both text and time-series data using shared embeddings. \cite{wang2024chattime} introduces ChatTime, a time-series foundation model that facilitates various zero-shot time-series tasks through continuous pretraining and instruction tuning on pretrained language models. \cite{li2025language} presents Texts as Time Series (TaTS), a multimodal time-series forecasting framework that incorporates concurrent textual data by converting it into auxiliary variables. This approach enables seamless integration of text-augmented time series into existing time-series models.

While prior work demonstrates the potential of LLMs for time-series forecasting, \textit{\textbf{none integrate them into state-space models—our key contribution.}}This integration enhances forecasting performance and enables principled uncertainty quantification.

% Time-MMD, HybridMMF(TTC)

\paragraph{Bayesian State Space Models.}

%Introduced by the seminal work of \cite{kalman1960new}, the classical Kalman filter was the first recursive Bayesian estimator that is optimal for linear dynamical systems with Gaussian noise. The abundance of non-linear systems in real-world scenarios soon motivated the extended Kalman filter~\citep{julier2004unscented} which makes linear approximations of functions to obtain state estimates. For more complex non-linear and non-Gaussian systems, sampling-based estimators such as the Unscented Kalman Filter~\citep{julier2004unscented}, Ensemble Kalman Filter~\citep{houtekamer1998data}, and sequential Monte Carlo methods~\citep{doucet2001sequential} that pass stochastic or deterministic samples through the system to estimate the posterior distribution have also been widely used in practice.

Bayesian state estimation has a long-standing history in control theory and time-series analysis. The classical Kalman filter~\cite{kalman1960new}, provides an optimal recursive solution for state estimation in linear dynamical systems with Gaussian noise. To accommodate the nonlinearities common in real-world systems, the Extended Kalman Filter was developed by linearizing nonlinear functions via Taylor expansion around the current estimate. Later, the Unscented Kalman Filter was introduced to improve upon EKF by using deterministic sampling to better capture the mean and covariance of nonlinear transformations~\cite{julier2004unscented}. Other sampling-based methods, such as the Ensemble Kalman Filter~\cite{houtekamer1998data} and Sequential Monte Carlo~\cite{doucet2001sequential}, have further advanced Bayesian filtering in nonlinear and non-Gaussian settings by representing posterior distributions through particle ensembles.

%More recently, there has been work across various fields and modalities integrating deep neural networks with Bayesian inference. \cite{(kvae)fraccaro2017disentangled} \cite{rangapuram2018deep} was the first to pioneer SSM parameter estimation for time-series forecasting. There has also been work modeling latent dynamics from pictures \cite{hafner2019learning}\cite{hafner2020mastering}\cite{revach2022kalmannet}. To the best of our knowledge, we are the first to explore the intersection between pretrained LLMs and Bayesian state estimation.

More recently, researchers have sought to combine the structure of state-space models with the flexibility of deep neural networks. For example, KVAE introduced variational approaches to learning latent dynamics in sequential data using neural parameterizations of the transition and emission functions~\cite{(kvae)fraccaro2017disentangled}. \cite{rangapuram2018deep} adapted state-space formulations for multivariate time-series forecasting in large-scale retail demand. In reinforcement learning and model-based control, works such as PlaNet~\cite{hafner2019learning}, Dreamer~\cite{hafner2020mastering}, and KalmanNet~\cite{revach2022kalmannet} have shown that combining deep neural networks with probabilistic latent dynamics models can yield strong performance across pixel-based partially observable domains.

Despite these advances, existing work largely targets unimodal data like images or numerical signals.
In contrast, \textit{\textbf{our work is the first to combine pretrained LLMs with probabilistic state-space models for joint forecasting over numeric and textual inputs, extending Bayesian state estimation to the multimodal setting.}}
\vfill

%\textbf{TODO: one paragraph on classical Bayesian filters and smoothers \citep{kalman1960new}\citep{kalman1961newresults}\citep{julier2004unscented}}

%\textbf{TODO: another paragraph on machine learning papers using Bayesian estimators \citep{(kvae)fraccaro2017disentangled}\citep{hafner2019learning}\citep{hafner2020mastering}\citep{revach2022kalmannet}\citep{rangapuram2018deep}}

%Originally studied in control and decision theory, Bayesian state estimators are used to infer the true state of a dynaThe seminal work by \cite{kalman1960new}, the Kalman filter, introduced an optimal recursive algorithm for estimating unobservable states of a dynamical system from noisy measurements. While initially developed for linear systems, following works such as the Unscented Kalman Filter (UKF) applicability to 

%Assuming a mathetmatical process over time, Bayesian state estimation, also known as Bayesian filtering, is a  Widely used in real-world appications on navigation systems, robotic motion control and planning, and econometric signal processing. Our work extends the applicability of Kalman filters to textual data by incorporating an autoencoder to filter state estimates based on pseudo-observations.

\section{Derivation of Training Objective }\label{appsec:training}

Our training objective in \autoref{eqn:objective} can be derived using the autoregressive structure of \ourmethod{} and Jensen's inequality.
\begin{align*}
    &\log p(\vy_{1:T}, \bm{\gD}_{1:T}) \\
    = &\log \int_{\vx_{1:T}}  p(\vy_{1:T}, \bm{\gD}_{1:T}, \vx_{1:T})d\vx_{1:T}\\
    = &\log \int_{\vx_{1:T}} \prod_{t=1}^T p(\vx_t \mid \vx_{t-1}) p(\vy_t \mid \vx_t) p(\bm{\gD}_t \mid \vx_t) d\vx_{1:T}\\
    = &\log \E_{q(\vx_{1:T} \mid \vy_{1:T},\bm{\gD}_{1:T})} \left[\dfrac{\prod_{t=1}^T p(\vx_t \mid \vx_{t-1}) p(\vy_t \mid \vx_t) p(\bm{\gD}_t \mid \vx_t)}{{q(\vx_{1:T} \mid \vy_{1:T}, \bm{\gD}_{1:T})}} \right]\\
    \geq & \;\E_{q(\vx_{1:T} \mid \vy_{1:T},\bm{\gD}_{1:T})} \left[ \log\dfrac{\prod_{t=1}^T p(\vx_t \mid \vx_{t-1}) p(\vy_t \mid \vx_t) p(\bm{\gD}_t \mid \vx_t)}{{q(\vx_{1:T} \mid \vy_{1:T}, \bm{\gD}_{1:T})}}\right]\\
    = & \E_{q(\vx_{1:T} \mid \vy_{1:T},\bm{\gD}_{1:T})} \left[\sum_{t=1}^t \log p(\vy_t \mid \vx_t)  + \log p(\bm{\gD}_t \mid \vx_t) + \log \dfrac{p(\vx_t \mid \vx_{t-1})}{{q(\vx_t \mid \vy_{1:t}, \bm{\gD}_{1:t})}}\right]\\
    = &\sum_{t=1}^T \E_{q(\vx_t \mid \vy_t,\bm{\gD}_t)} \bigg[\underbrace{\log p(\vy_t \mid \vx_t)}_{\text{value likelihood}} + \underbrace{\log p(\bm{\gD}_t \mid \vx_t)}_{\text{text likelihood}}\bigg] - \underbrace{\text{KL}(q(\vx_t \mid \vy_{1:t},\bm{\gD}_{1:t}) \mid\mid p(\vx_t \mid \vx_{t-1}))}_{\text{temporal regularization}} \label{eqn:objective}
\end{align*}

%While integrating an LLM with SSM enables new capabilities, it also presents unique challenges in implementation. Therefore we close the section by discussing details of the training pipeline that enable tractable and effective training of \ourmethod{}.

%\paragraph{Stateful single-step training.} Ideally, training \ourmethod{} on long sequences would allow the model to better capture long-range dependencies. However, each timestep $t$ requires passing $\bm{\gD}_t$ through the LLM twice---once for encoding and once for decoding---making na\"ive long-horizon training computationally expensive. To address this, we adopt \textit{stateful} training~\cite{yen2022stateful,katrompas2022enhancing}, where the model is trained on single-step batches while preserving its temporal ordering, with hidden states passed onto the next training iteration. Starting from a fixed prior $\vx_0 = 0$ (and hidden $\vh_0 = 0$), the training process is illustrated in Algorithm~\ref{alg:training_step}.
\section{Additional Experiment Details}\label{appsec:extended_results}

\paragraph{Datasets.} \textsc{TTC-Climate} is consisted of daily temperature measurements at Washington DC, accompanied by textual weather descriptions spanning from January 1st, 2014 to December 1st, 2023. \textsc{TTC-Medical} stores daily heart rate measurements from 73 patients accompanied by nursing notes writing observations and treatment plans. Each patient data spans an average length of 104 days. Following previous work~\cite{kim2024multi}, we train the model on the first 80\% of all timestamps, validate on the next 10\%, then test on the last 10\%. 

\paragraph{LLM Prompts.} For both \textsc{TTC-Climate} and \textsc{TTC-Medical} experiments, we use the following prompts for text-conditioned posterior estimation and state-conditioned text generation, respectively.

\begin{figure}[!h]
    \centering
    \begin{tcolorbox}[title={Prompt for text-conditioned posterior estimation}]
    User: Encode the information into a sequence of vectors. \texttt{<INSERT TEXT>} \\
    Assistant: \texttt{<INSERT SUMMARY TOKENS>}
    \end{tcolorbox}
    \vspace{-5px}
    \begin{tcolorbox}[title={Prompt for state-conditioned text generation}]
    User: \texttt{<INSERT STATE>} Given this belief state, generate a textual forecast.\\ Date: \texttt{<INSERT FORECAST DATE AS YYYY-MM-DD>}\\
    Assistant: 
    \end{tcolorbox}
    %\vspace{-5px}
    %\caption{Prompts}
    %\label{fig:ttc_prompts}
\end{figure}

\paragraph{Models.} For the SSM backbone, we use a single layer GRU with state and hidden dimensions both equal to 16. For the LLM, we use LLaMA3.1-8B~\cite{grattafiori2024llama} as the default model, and adapt the MLP weights is all layers using LoRA with rank and alpha parameters equal to 8 and 16, respectively. For text compression, we augment and use a set of 8 summary tokens. Similarly for textual forecasting, we project the states into 8 prefix tokens, which are prepended for in-context generation.

\newpage
\paragraph{Optimization.}  For all experiments, we use the AdamW optimizer with a learning rate that follows a cosine annealing schedule, starting from 5e-4 and reduced towards 5e-5 during training. We run a maximum of 20 training epochs, and if the model does not improve its validation loss for 5 consecutive epochs, we stop early to prevent further overfitting. Following previous work~\cite{hafner2019learning}, we use a free nats parmaeter set to 2.5, which effectively clamps the KL loss and thus allows the model to learn meaningful latents at the beginning of training. This free nats parameters is linearly annealed towards zero during training.

Our training process uses the AdamW optimizer in combination with a cosine decay schedule that initiates at a learning rate of 5e-4 and anneals gradually to 5e-5. Each model is trained for up to 20 epochs, with early termination triggered if validation performance fails to improve over five successive epochs. Inspired by strategies in prior latent sequence modeling~\cite{hafner2019learning}, we introduce a "free nats" threshold of 2.5 to restrict the KL penalty early in training. This constraint encourages the model to utilize its latent capacity more effectively at initialization and is gradually reduced to zero as optimization proceeds.

\paragraph{Loss weighting.} Despite using a small LoRA rank, the number of trainable parameters in the LLM still far exceeds those in the SSM. Consequently, we find that uniform weighting of the loss components in Equation~\ref{eqn:objective} tends to bias optimization toward the text likelihood term, often overfitting to language modeling while underfitting on structured numerical predictions. Although dynamic or adaptive weighting schemes (e.g., uncertainty-based or gradient norm balancing) could be employed~\cite{kendall2018multi}, we find that a simple weighting scheme with $\alpha_{\text{val}} = \alpha_{\text{KL}} = 1.0$ and $\alpha_{\text{text}} = 0.1$ provide a good trade-off between tasks without requiring additional tuning.

\paragraph{Full results.}

\autoref{tab:ttc_combined_main_results} shares the full results on \textsc{TTC-Climate} and \textsc{TTC-Medical}, using all unimodal and multimodal methods considered in our work.

\begin{table}[!h]
    \centering
    \begin{subtable}{\textwidth}
    \centering
    \begin{tabular}{c|ccccccc}
        \toprule
        Method & $H = 1$ & 2 & 3 & 4 & 5 & 6 & 7 \\
        \midrule
        PatchTST~\cite{(patchtst)nie2022time} & 4.912 & 5.305 & 6.021 & 6.576 & 6.980 & 7.170 & 7.360\\
        NLinear~\cite{(nlinear)zeng2023transformers} & 4.981 & 6.129 & 6.501 & 6.710 & 6.834 & 6.916 & 6.962\\
        NLinear-Text~\cite{(nlinear)zeng2023transformers} & 4.835 & 5.800 & 5.951 & 5.934 & 6.022 & 6.024 & 6.106 \\
        TSFLib~\cite{liu2024time} & 6.351 & 6.446 & 6.143 & 6.360 & 6.096 & 6.379 & \bf 6.002\\
        TT2TT~\cite{kim2024multi} & 5.243 & 5.955 & 6.724 & 7.253 & 7.678 & 8.034 & 7.666\\
        HybridMMF~\cite{kim2024multi} & 4.759 & 5.597 & 5.906 & 6.019 & 6.133 & 6.027 & 6.143\\
        \midrule
        \ourmethod{} (Unimodal) & 4.224 & 5.029 & 5.523 & 5.855 & 6.107 & 6.303 & 6.473 \\
        \ourmethod{} (Multimodal) & \bf 4.117 & \bf 4.908 & \bf 5.341 & \bf 5.627 & \bf 5.833 & \bf 5.998 & 6.133 \\
        \bottomrule
    \end{tabular}
    \caption{\textsc{TTC-Climate}}
    \end{subtable}
    \begin{subtable}{\textwidth}
    \centering
    \begin{tabular}{c|ccccccc}
        \toprule
        Method & $H = 1$ & 2 & 3 & 4 & 5 & 6 \\
        \midrule
        PatchTST~\cite{(patchtst)nie2022time} & 5.735 & 6.757 & 7.350 & 7.687 & 7.996 & 8.470\\
        NLinear~\cite{(nlinear)zeng2023transformers} & 5.195 & 5.279 & 5.275 & 5.406 & 5.562 & 5.875\\
        NLinear-Text~\cite{(nlinear)zeng2023transformers} & 5.117 & 5.143 & 5.106 & 5.300 & 5.492 & 5.759 \\
        TSFLib~\cite{liu2024time} & 6.767 & 7.066 & 7.427 & 7.050 & 7.165 & 7.210 \\
        TT2TT~\cite{kim2024multi} & 6.689 & 6.432 & 6.022 & 6.483 & 6.747 & 6.731 \\
        HybridMMF~\cite{kim2024multi} & 5.202 & 5.472 & 6.620 & 6.269 & 8.673 & 8.454 \\
        \midrule
        \ourmethod{} (Unimodal) & 3.910 & 4.598 & 5.047 & 5.268 & 5.473 & 5.654\\
        \ourmethod{} (Multimodal) & \bf 3.583 & \bf 4.268 & \bf 4.721 & \bf 5.043 & \bf 5.296 & \bf 5.487\\
        \bottomrule
    \end{tabular}
    \caption{\textsc{TTC-Medical}}
    \end{subtable}
    %\vspace{5px}
    \caption{Test RMSE results from \textsc{TTC} benchmark. Best results for each prediction horizon $H$ are highlighted in \textbf{bold}.}
    \label{tab:ttc_combined_main_results}
\end{table}

% Don't forget to add back before submission

\end{document}